\documentclass[lettersize,journal]{IEEEtran}
\usepackage{amsmath,amsfonts}
\usepackage{algorithmic}
\usepackage{algorithm}
\usepackage{array}
\usepackage[caption=false,font=normalsize,labelfont=sf,textfont=sf]{subfig}
\usepackage{textcomp}
\usepackage{stfloats}
\usepackage{url}
\usepackage{verbatim}
\usepackage{graphicx}
\usepackage{cite}
\usepackage{amssymb}
\usepackage{booktabs}
\usepackage{mathtools}
\begin{document}

\title{Robust Learning under Non-Gaussian Noise with Adaptive Log-Correntropy Loss}

\author{Mainak Kundu, Ria Kanjilal, and Ismail Uysal
\thanks{Mainak Kundu and Ismail Uysal are with the Department of Electrical Engineering, University of South Florida, Tampa, FL 33620 USA (e-mail: mkundu@usf.edu, iuysal@usf.edu).}
\thanks{Ria Kanjilal is with the Department of Computer Engineering, California Polytechnic State University, San Luis Obispo, CA 93407 USA (e-mail: rkanjila@calpoly.edu)}}

\maketitle

\begin{abstract}
Robust deep learning under heavy-tailed and impulsive noise remains challenging because conventional losses such as mean squared error (MSE) exhibit unbounded sensitivity to outliers. Although correntropy-based objectives improve robustness, many conventional formulations rely on fixed kernel parameters that must be empirically tuned and remain static during training. To address these limitations, we propose an Adaptive Log-Correntropy Loss (ALCL), a heavy-tailed loss formulation that adaptively learns its robustness geometry during optimization. ALCL introduces a logarithmic residual model whose shape and scale parameters are learned jointly with network weights through differentiable reparameterization. This yields a heavy-tailed robust optimization objective whose $\psi$-function is formally bounded and redescending, allowing the loss geometry to adapt dynamically to evolving residual statistics while suppressing extreme outliers. Comparative experiments on four widely used benchmark datasets spanning grayscale and red–green–blue (RGB) image data under mixed heavy-tailed and impulsive noise demonstrate that ALCL consistently outperforms MSE and optimally tuned generalized correntropy losses in both reconstruction fidelity and downstream classification accuracy. While performance differences remain small under low-noise conditions, under high-noise regimes ALCL improves median accuracy by up to 4.75\% on grayscale benchmarks and 4.51\% on RGB datasets, with reduced variance across runs. These results demonstrate that adaptive robustness through joint learning of loss parameters provides a computationally efficient alternative to static correntropy-based losses for deep learning in non-Gaussian environments.
\end{abstract}

\begin{IEEEkeywords}
Adaptive Log-Correntropy Loss, Information-theoretic learning, Robust deep learning, Non-Gaussian noise, Joint optimization, Maximum correntropy criterion
\end{IEEEkeywords}

\section{Introduction}
\label{sec1}

\IEEEPARstart{R}{ecent} aadvances in deep learning have enabled the widespread adoption of data-driven models in critical domains such as medical imaging, computer vision, surveillance, and autonomous systems. In these real-world settings, the loss function defines the optimization objective by specifying how prediction errors are measured and how gradients are formed during backpropagation, directly influencing convergence behavior, sensitivity to outliers, and overall model accuracy~\cite{goodfellow2016deep}. A commonly used objective for reconstruction and regression tasks is mean squared error (MSE), which is consistent with a Gaussian residual model under a maximum-likelihood interpretation. However, Gaussian residual assumptions are frequently violated in practice by impulsive interference and heavy-tailed outliers~\cite{Shao1993,chen2016efficient}. Under such conditions, the quadratic growth of MSE assigns unbounded influence to large residuals, allowing outliers to dominate gradient updates and distort learned representations. Information-theoretic learning (ITL) addresses this limitation through the maximum correntropy criterion (MCC), which replaces global error energy with a localized kernel-based similarity measure~\cite{principe2010information}. While correntropy-based objectives offer improved resilience to non-Gaussian noise~\cite{qi2014robust}, their robustness characteristics are typically static, with kernel shape and scale treated as fixed or empirically selected hyperparameters~\cite{chen2016efficient}. As a result, the robustness profile of the loss is determined a priori and may become mismatched to the evolving residual distribution during training. Although heuristic adaptation strategies have been explored~\cite{ge2024gaussian}, some approaches employ separate parameter-update procedures from the main optimization process and do not provide a principled mechanism for jointly learning robustness parameters with network weights. These limitations motivate the development of loss formulations whose robustness properties can evolve during gradient-based optimization.

In practical sensing and imaging systems, departures from Gaussian noise can arise from sensor imperfections, measurement errors, transmission disturbances, and impulsive interference. Such corruption is often characterized by heavy-tailed behavior, in which large deviations occur more
frequently than predicted by a Gaussian model and cannot be adequately described using only second-order statistics~\cite{principe2010information}. This has motivated the use of information-theoretic learning criteria, notably minimum error entropy (MEE) and the maximum correntropy criterion (MCC), for learning under non-Gaussian and heavy-tailed conditions ~\cite{principe2000information,erdogmus2002error,wang2021numerically}. MEE exploits information from the error distribution beyond conventional second-order measures but generally involves greater computational complexity due to its pairwise error interactions. MCC, in contrast, avoids the pairwise error interactions required by conventional MEE formulations, offering a computationally attractive alternative for robust learning and adaptive filtering~\cite{principe2010information,zhao2023efficient}. 

To address these limitations, we propose Adaptive Log-Correntropy Loss (ALCL), a shape- and scale-adaptive extension of the Cauchy/Lorentzian robust loss for representation learning under impulsive and heavy-tailed noise. Unlike conventional robust objectives whose robustness characteristics are fixed before training, ALCL incorporates shape and scale as trainable loss parameters and optimizes them within the same training process as the network parameters, allowing the error-response characteristics of the loss to evolve during optimization. From a robust estimation perspective, we derive the $\psi$-function of ALCL and establish its bounded and redescending properties, showing that the proposed loss progressively suppresses the contribution of extreme residuals while retaining sensitivity to informative errors. To support stable optimization, we further develop a differentiable parameterization framework that incorporates the adaptive robustness parameters directly into the backpropagation process. Comprehensive experiments on grayscale and RGB benchmarks under different levels of heavy-tailed and impulsive corruption demonstrate consistent improvements in downstream classification accuracy over MSE and carefully tuned GGCL, supported by F1-score comparisons and convergence analysis. The main contributions of this paper are summarized as follows: (i) We formulate ALCL as a shape- and scale-adaptive logarithmic robust loss for deep learning, enabling its robustness characteristics to evolve through jointly learned loss parameters. (ii) We derive the residual influence ($\psi$) function of ALCL and establish its bounded and redescending properties, providing a mathematical explanation of its robustness against extreme residuals. (iii) We develop a differentiable parameterization framework for the shape and scale parameters, enabling their joint gradient-based optimization with network
weights during backpropagation. (iv) We demonstrate improved downstream classification accuracy
across all evaluated noise conditions and higher F1-scores under high non-Gaussian noise, while convergence analysis reveals sustained reductions in normalized training and validation losses compared with the earlier stabilization observed for the fixed-loss baselines.

To ensure a fair comparison, we evaluate ALCL on four benchmark datasets using identical network architectures and competing loss functions.The rest of the paper is organized as follows. Section~\ref{sec2} reviews related work on correntropy-based learning and robust loss functions. Section~\ref{sec3} introduces the proposed ALCL formulation, including its probabilistic interpretation and robustness analysis, followed by Section~\ref{sec4}, which presents ALCL-based autoencoder architectures for both grayscale and RGB data. Section~\ref{sec5} describes the experimental setup and noise models, while Section~\ref{sec6} reports quantitative and qualitative results. Finally, Section~\ref{sec7} concludes the paper and outlines future research directions.

\section{Related work}\label{sec2}

Robust loss-function design has evolved from conventional error-based criteria toward information-theoretic approaches capable of handling non-Gaussian and impulsive disturbances. MSE remains one of the most widely used objectives because of its simple optimization characteristics and its connection to Gaussian error modeling. However, its quadratic growth makes it highly sensitive to large residuals. Absolute-error-based objectives such as the mean absolute error (MAE) reduce this sensitivity through a linear penalty, while the Huber loss combines quadratic behavior for small residuals with a linear penalty for larger errors, providing a classical compromise between estimation efficiency and robustness to outliers~\cite{hampel1992introduction}. Despite these improvements, such point wise error-based losses employ a predetermined penalty geometry and do not explicitly exploit higher-order information contained in the error distribution. This limitation has motivated the development of information-theoretic learning criteria for environments in which Gaussian assumptions are inadequate.

Within information-theoretic learning, correntropy emerged as a localized similarity measure capable of exploiting higher-order statistical information and was subsequently adopted through the maximum correntropy criterion (MCC) for robust learning under non-Gaussian disturbances ~\cite{liu2006correntropy,principe2010information}. Subsequent research has extended the original Gaussian-kernel formulation in several directions,including its integration into robust representation learning, generalized kernel formulations, logarithmic correntropy criteria, heavy-tailed kernels,mixture and multi-kernel approaches, and adaptive kernel strategies. The
following subsections review this progression and the remaining limitations that motivate the proposed ALCL framework.

\subsection{Correntropy and maximum correntropy criterion}

Correntropy is a localized similarity measure between two random variables $X$ and $Y$, defined as ~\cite{liu2006correntropy}

\begin{equation}
        V_\sigma(X,Y) = \mathbb{E}\left[\kappa_\sigma(X - Y)\right],
\label{eq1}
\end{equation}

\noindent where $e = X - Y$ is the residual (error) between the two random variables, $\mathbb{E}$ denotes the mathematical expectation and $\kappa_\sigma(\cdot)$ is the Gaussian kernel with $\sigma$ as the kernel size:

\begin{equation}
        \kappa_\sigma(e) = \frac{1}{\sqrt{2\pi}\sigma} \exp\left(-\frac{e^2}{2\sigma^2}\right).
\label{eq2}
\end{equation}

\noindent In practical learning scenarios, correntropy is estimated using finite samples 

\begin{equation}
        \hat{V}_\sigma(X,Y) = \frac{1}{N} \sum_{i=1}^{N} \kappa_\sigma(x_i - y_i),
\label{eq3}
\end{equation}

\noindent where $X$ and $Y$ denote the two random variables of interest, and $\{(x_i, y_i)\}_{i=1}^{N}$ are $N$ paired samples drawn from their joint distribution.

The maximum correntropy criterion optimizes model parameters by maximizing correntropy rather than minimizing squared error, yielding strong robustness under impulsive and non-Gaussian noise~\cite{wang2015regularized}. However, MCC is highly sensitive to the kernel width $\sigma$, motivating adaptive kernel strategies that update $\sigma$ based on instantaneous error statistics~\cite{zhao2012adaptive,radmanesh2017recursive,wang2015adaptive}, commonly using rules as

\begin{equation}
        \sigma_n^2 = \sigma_0^2 + e_n^2 .
\label{eq4}
\end{equation}

\noindent While effective in adaptive filtering, such updates are typically defined through separate error-dependent rules rather than incorporated as trainable loss parameters during network optimization.

\subsection{Correntropy in deep learning}

The robustness of correntropy subsequently motivated its integration into neural representation learning. Qi et al.~\cite{qi2014robust} incorporated the maximum correntropy criterion into stacked autoencoders as an alternative to
MSE, demonstrating improved feature learning in the presence of non-Gaussian noise and outliers. Chen et al.~\cite{chen2016efficient} further investigated correntropy-induced loss for robust stacked autoencoders, showing 
that correntropy-based reconstruction can reduce the influence of corrupted observations during feature extraction. These studies established correntropy as a useful robust objective for deep representation learning; however, their
robustness behavior remained dependent on predefined kernel parameters.
\begin{equation}
        J_{\text{MCC}}(\theta) = \frac{1}{MN} \sum_{i=1}^{M} \sum_{j=1}^{N} \kappa_\sigma(x_{ij} - \hat{x}_{ij}),
\label{eq5}
\end{equation}

\noindent where $x_{ij}$ and $\hat{x}_{ij}$ denote the $j$-th feature of the $i$-th input and reconstructed samples, respectively. In \eqref{eq5}, $M$ represents the number of samples and $N$ denotes the feature dimensionality of each sample. Recent findings further indicate that robust losses such as correntropy play a critical role in stabilizing feature extraction layers, yielding more resilient representations in multi-stage learning pipelines~\cite{liu2023robust}.

\subsection{Generalized and logarithmic correntropy loss functions}

To overcome the limitations of Gaussian correntropy, generalized correntropy was introduced using a generalized Gaussian density kernel denoted as~\cite{hu2018diffusion}

\begin{equation}
    \begin{aligned}
        G_{\alpha,\beta}(e)
            &= \frac{\alpha}{2\beta\,\Gamma(1/\alpha)}\exp\!\left(-\left|\frac{e}{\beta}\right|^{\alpha}\right) \\
            &= \gamma_{\alpha,\beta}\exp\!\left(-\lambda |e|^{\alpha}\right),
    \end{aligned}
\label{eq6}
\end{equation}

\noindent where $\alpha > 0$ denotes the shape parameter, $\beta > 0$ denotes the bandwidth (scale) parameter, and $\Gamma(\cdot)$ is the Gamma function. In \eqref{eq6}, $\lambda = 1/\beta^{\alpha}$ represents the kernel parameter and $\gamma_{\alpha,\beta} = \alpha / \left(2\beta \Gamma(1/\alpha)\right)$ is the normalization constant. When $\alpha \rightarrow \infty$, the generalized Gaussian distribution converges pointwise to a uniform distribution on the interval $(-\beta, \beta)$. The Laplacian distribution ($\alpha = 1$) and the Gaussian distribution ($\alpha = 2$) arise as special cases of this parametric family of symmetric distributions. Based on generalized correntropy, the generalized correntropy loss (GC-loss) between two random variables $X$ and $Y$ is given by~\cite{chen2016generalized}

\begin{equation}
        J_{\text{GC-loss}}(X,Y) = G_{\alpha,\beta}(0) - V_{\alpha,\beta}(X,Y),
\label{eq7}
\end{equation}

\noindent where $G_{\alpha,\beta}(\cdot)$ denotes the generalized Gaussian kernel with shape parameter $\alpha>0$ and scale parameter $\beta>0$, and
$V_{\alpha,\beta}(X,Y)=\mathbb{E}[G_{\alpha,\beta}(X-Y)]$ is the generalized correntropy. In practice, given $N$ paired samples $\{(x_i,y_i)\}_{i=1}^{N}$, the GC-loss is estimated as

\begin{align}
\hat{J}_{\mathrm{GC\text{-}loss}}(X,Y)
&= G_{\alpha,\beta}(0) - \hat{V}_{\alpha,\beta}(X,Y) \\
&= \gamma_{\alpha,\beta} - \frac{1}{N} \sum_{i=1}^{N} G_{\alpha,\beta}(e_i),
\label{eq8-9}
\end{align}

\noindent where $e_i = x_i - y_i$ represents the residual of the $i$-th sample and $\gamma_{\alpha,\beta}=G_{\alpha,\beta}(0)$ is the normalization constant. To improve robustness under non-stationary and impulsive noise, logarithmic transformations have been incorporated into correntropy-based learning frameworks by applying a logarithmic modulation to the correntropy-induced error cost. In~\cite{hu2018diffusion}, the resulting logarithmic correntropy loss is defined as

\begin{equation}
V_{\sigma}(X,Y)
= \mathbb{E}\!\left[
\kappa_{\sigma}(X-Y)
- \frac{1}{\delta}\,
\ln\!\left(1+\delta\,\kappa_{\sigma}(X-Y)\right)
\right],
\label{eq10}
\end{equation}

\noindent where $\delta>0$ is a fixed system parameter that controls the degree of logarithmic compression and the corresponding sample estimator follows as

\begin{equation}
\hat{V}_{\sigma}(X,Y)
= \frac{1}{N}\sum_{i=1}^{N}
\left[
\kappa_{\sigma}(e_i)
- \frac{1}{\delta}\ln\!\left(1+\delta\,\kappa_{\sigma}(e_i)\right)
\right].
\label{eq11}
\end{equation}

\subsection{Cauchy kernel-based correntropy}

Recent research has introduced the Cauchy kernel as a robust alternative to the standard Gaussian kernel in non-Gaussian signal processing \cite{liu2025cauchy}. The Cauchy kernel is defined as

\begin{equation}
    C_{\delta}(e) = \frac{1}{1 + e^2/\delta}
\label{eq12}
\end{equation}

\noindent where $\delta$ is the kernel bandwidth. Unlike the exponential decay of Gaussian functions, the Cauchy kernel’s polynomial decay ensures the maximum cauchy correntropy criterion is less sensitive to extreme outliers. A key theoretical advantage is its bounded sensitivity to the bandwidth parameter $\delta$

\begin{equation}
    \sup_{e \in \mathbb{R}} \left| \frac{\partial C_{\delta}}{\partial \delta} \right| = \sup_{e} \left| \frac{e^2}{\delta^2(1 + e^2/\delta)^2} \right| = \frac{1}{4\delta}.
\label{eq13}
\end{equation}

\noindent This property prevents numerical instability and singular matrix issues common in traditional correntropy filters, providing a stable foundation for robust estimation in impulsive noise environments. However, despite these advantages, Cauchy kernel–based correntropy remains a fixed-kernel formulation, with robustness governed by a predefined kernel shape and bandwidth.

\subsection{Mixture correntropy and multi-kernel approaches}

To reduce bandwidth sensitivity in single-kernel correntropy, mixture correntropy combines multiple kernels to simultaneously handle small and large residuals~\cite{luo2021towards, zheng2020mixture}. A generic mixture correntropy kernel is represented as

\begin{equation}
        \kappa_{\text{mix}}(e) = \alpha \kappa_{\sigma_1}(e) + (1-\alpha)\kappa_{\sigma_2}(e),
\label{eq14}
\end{equation}

\noindent where $\alpha \in [0,1]$ controls the contribution of each kernel component, $\kappa_{\sigma_1}(\cdot)$ and $\kappa_{\sigma_2}(\cdot)$ denote kernels of the same functional form but with different kernel sizes, and $e = x - y$ is the residual. This formulation underpins the maximum mixture correntropy criterion, which utilizes a sample estimator \cite{zheng2020mixture}

\begin{equation}
        \hat{V}(X,Y) = \frac{1}{N}\sum_{j=1}^{N} \left[ \alpha G_{\sigma_1}(x_j,y_j) + (1-\alpha) G_{\sigma_2}(x_j,y_j) \right].
\label{eq15}
\end{equation}

\noindent In \eqref{eq15}, $\{x_j\}_{j=1}^N$ and $\{y_j\}_{j=1}^N$ are paired samples, where $N$ denotes the number of samples, $G_{\sigma_s}(\cdot,\cdot)$ is the Gaussian correntropy kernel, and $\sigma_1, \sigma_2$ correspond to distinct scales. This concept has been extended to heterogeneous kernel designs, most notably the Gaussian--Cauchy mixture, which explicitly combines light- and heavy-tailed behaviors as ~\cite{ge2024gaussian}

\begin{equation}
\kappa_{\text{GC}}(e)
=
\alpha \exp\!\left(-\frac{e^2}{2\sigma^2}\right)
+
(1-\alpha)\frac{1}{1+\frac{e^2}{c\sigma}},
\label{eq16}
\end{equation}

\noindent where $\sigma$ is the scale parameter and $c$ is the heavy-tail scaling factor of the Cauchy component, jointly controlling the degree of outlier suppression.

\subsection{Adaptive correntropy and variable kernel strategies}

Tan et al.~\cite{tan2012adaptive} addressed the sensitivity of correntropy to kernel-size selection by deriving an adaptive kernel-size estimation strategy based on the maximum correntropy criterion. Their method determines the kernel parameter from feature-distribution separability and was demonstrated for correntropy-based face-recognition metrics.Subsequent studies developed adaptive kernel-width strategies in which the bandwidth is adjusted according to the observed error characteristics \cite{zhao2011kernel,wang2018kernel}. Lin et al.~\cite{lin2022maximum} further extended adaptive correntropy to multidimensional signals by introducing a variable kernel-width strategy within the maximum total quaternion correntropy framework. More recently, Nguyen et al.~\cite{nguyen2026robust} proposed an adaptive multi-kernel mixture correntropy framework in which kernel centers, kernel sizes, and mixture coefficients are determined from the empirical error distribution, reducing dependence on manually selected free parameters.

These studies demonstrate the growing importance of adapting correntropy parameters to the observed error characteristics. However, such adaptation is typically realized through dedicated parameter-selection rules, error-distribution estimation, or auxiliary update procedures. In contrast,ALCL treats its shape and scale parameters as trainable components of the loss and optimizes them together with the network parameters during backpropagation, thereby integrating robustness adaptation directly into the deep-learning optimization process.

\section{Proposed ALCL framework}\label{sec3}

Building on the limitations of fixed and adaptively selected correntropy kernels discussed in Section~\ref{sec2}, we propose Adaptive Log-Correntropy Loss (ALCL), a shape- and scale-adaptive generalization of the Cauchy/Lorentzian robust penalty~\cite{black1996unification}. Instead, ALCL defines its robustness characteristics directly in the residual space through a logarithmic heavy-tailed penalty with learnable shape and scale parameters.These parameters are optimized jointly with the network weights through gradient-based learning, allowing the loss geometry to adapt dynamically during training rather than remaining constrained by a fixed kernel family or fixed loss parameters.

\subsection{Robust residual modeling and ALCL formulation}

Let $\mathbf{x} \in \mathbb{R}^d$ be the original clean input feature vector, where $d$ is the dimensionality of the output space, and let $\hat{\mathbf{x}} = f_\theta(\cdot)$ denote the network output. The reconstruction residual vector is defined as

\begin{equation}
    \mathbf{e} = \mathbf{x} - \hat{\mathbf{x}}.
\label{eq18}
\end{equation}

\begin{figure}[t]
    \centering
    \includegraphics[width=\linewidth]{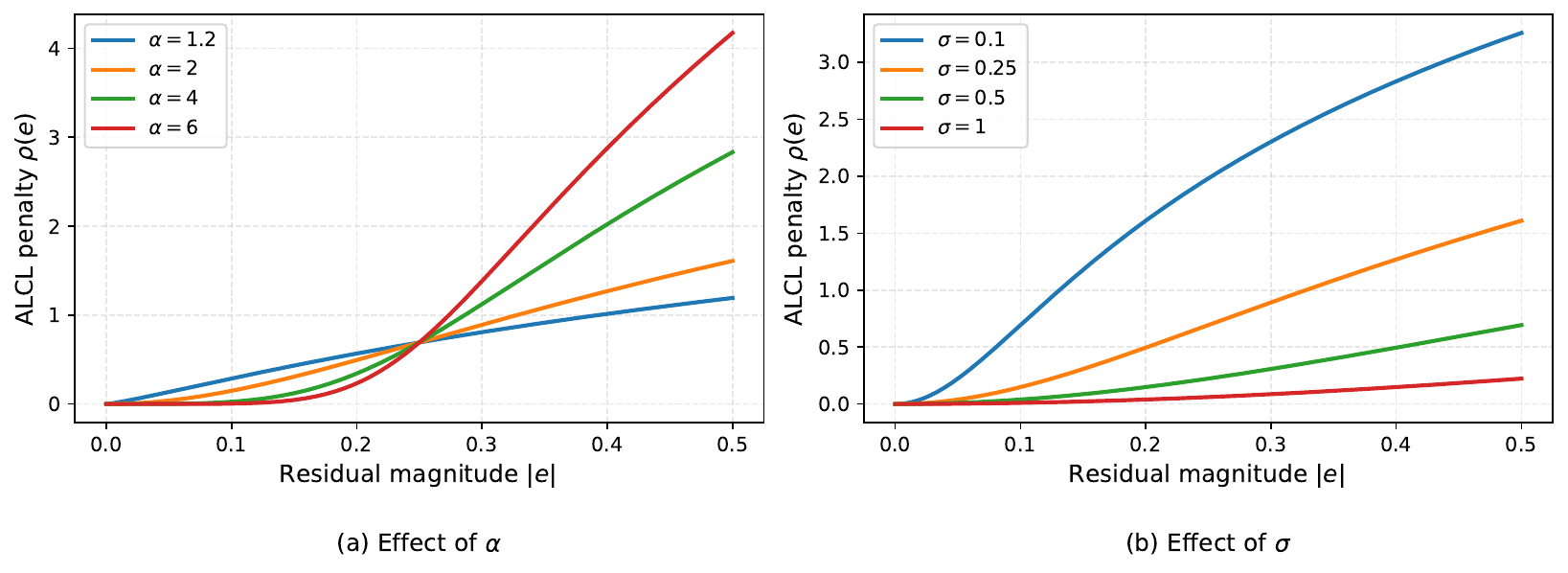}
    \caption{Effect of the shape and scale parameters on the ALCL penalty as a function of reconstruction residual magnitude.(a) Variation in loss curvature for different values of $\alpha$ with fixed $\sigma=0.25$. (b) Variation in residual penalization for different values of $\sigma$ with fixed $\alpha=2$.The curves illustrate how the two parameters control the error-response characteristics of ALCL.}
    \label{fig:ALCL_shape_scale_analysis}
\end{figure}

\noindent Under heavy-tailed or impulsive noise, quadratic losses impose unbounded influence from large residuals, leading to unstable training. ALCL addresses this through a logarithmic heavy-tailed penalty defined directly on the reconstruction residual. For a single residual component $e$, the loss is defined as

\begin{equation}
    \mathcal{L}_{\text{ALCL}}(e;\alpha,\sigma) = \ln\!\left(1 + \left(\frac{|e|}{\sigma}\right)^{\alpha}\right),
\label{eq19}
\end{equation}
\medskip

\noindent where $\alpha>1$ and $\sigma>0$ denote the adaptive shape and scale parameters, respectively. The shape parameter controls the curvature and relative penalization of residuals,
while the scale parameter determines the characteristic residual magnitude separating the near-origin and heavy-tailed regimes. Specifically, for $|e|\ll\sigma$, the ALCL penalty behaves approximately as $(|e|/\sigma)^\alpha$, whereas for $|e|\gg\sigma$, it approaches $\alpha\ln(|e|/\sigma)$.
The logarithmic structure ensures diminishing gradient contributions from extreme residuals. However, these parameter constraints alone do not guarantee effective optimization, as the gradient response also depends on the relative magnitudes of the residuals and the scale parameter. Consequently, appropriate parameter initialization and adaptation are important for reconstruction learning. 
Figure~\ref{fig:ALCL_shape_scale_analysis} illustrates how the shape and scale parameters influence the ALCL penalty across different reconstruction residual magnitudes.

\subsection{Joint learning of robustness parameters}

The ALCL penalty can be associated with a heavy-tailed error model whose
unnormalized density has the form

\begin{equation}
    p(e \mid \alpha,\sigma)\propto
    \left(1 + \left(\frac{|e|}{\sigma}\right)^{\alpha}\right)^{-1}.
\label{eq20}
\end{equation}

\noindent Motivated by this heavy-tailed formulation, the proposed learning
objective jointly optimizes the network parameters and the shape and scale
parameters as

\begin{equation}
    \min_{\theta,\alpha,\sigma} \mathbb{E}
    \left[
    \ln\!\left(1 + \left(\frac{|e|}{\sigma}\right)^{\alpha}\right)
    \right].
\label{eq21}
\end{equation}

\noindent Equation~\eqref{eq20} motivates the logarithmic heavy-tailed form of the proposed loss. During training, the ALCL objective in \eqref{eq21} is optimized directly with respect to the network parameters $\theta$ and the loss parameters $\alpha$ and $\sigma$. Thus, $\alpha$ and $\sigma$ are learned jointly with the network through gradient-based optimization rather than being fixed or tuned independently.

\subsection{Adaptive shape and scale parameterization}

To enable joint gradient-based optimization via backpropagation, the shape and scale parameters are learned through differentiable reparameterization. Unconstrained trainable variables $\phi,\psi\in\mathbb{R}$ are mapped to their admissible domains. The shape parameter is constrained to $(1,\infty)$ as

\begin{equation}
\alpha=f_{\mathrm{pos}}(\phi)+1,
\end{equation}

\noindent where $f_{\mathrm{pos}}(\cdot)$ denotes a smooth strictly positive mapping implemented using the softplus function $f_{\mathrm{pos}}(x)=\ln(1+e^x)$. Similarly, the scale parameter is mapped to the positive real axis as $\tilde{\sigma}=f_{\mathrm{pos}}(\psi)$, and the effective scale $\sigma$ is subsequently constrained to $[\sigma_{\min},\sigma_{\max}]$ to avoid degenerate scale values during optimization. The importance of adapting the scale parameter has been recognized in correntropy-based learning, where kernel size influences the weighting of residuals and the resulting robustness characteristics \cite{tan2012adaptive,wang2023theoretical}. Unlike kernel-size selection or error-dependent update strategies, ALCL treats both shape and scale as trainable parameters and optimizes them jointly with the network parameters through the reconstruction objective.

Specifically, let $\theta$ denote the network parameters and $\phi$ and $\psi$ the unconstrained variables associated with $\alpha$ and $\sigma$, respectively. For a mini-batch objective $\mathcal{L}_{\mathrm{ALCL}}$, the gradient-based updates are

\begin{align}
\theta^{(t+1)}
&=\theta^{(t)}
-\eta_{\theta}
\frac{\partial\mathcal{L}_{\mathrm{ALCL}}}
{\partial\theta^{(t)}},
\\
\phi^{(t+1)}
&=\phi^{(t)}
-\eta_{\alpha}
\frac{\partial\mathcal{L}_{\mathrm{ALCL}}}
{\partial\phi^{(t)}},
\\
\psi^{(t+1)}
&=\psi^{(t)}
-\eta_{\sigma}
\frac{\partial\mathcal{L}_{\mathrm{ALCL}}}
{\partial\psi^{(t)}},
\label{eq22-23}
\end{align}

\noindent where $\eta_{\theta}$, $\eta_{\alpha}$, and $\eta_{\sigma}$ denote the corresponding optimization step sizes. As illustrated in Fig.~\ref{fig:ALCL_flowchart}, because $\alpha$ and $\sigma$ enter the same reconstruction objective used to update $\theta$, the network representation and loss geometry are optimized jointly during training. The gradients of the reparameterized variables follow from the chain rule as

\begin{equation}
\frac{\partial\mathcal{L}_{\mathrm{ALCL}}}{\partial\phi}
=
\frac{\partial\mathcal{L}_{\mathrm{ALCL}}}{\partial\alpha}
\frac{\partial\alpha}{\partial\phi},
\qquad
\frac{\partial\mathcal{L}_{\mathrm{ALCL}}}{\partial\psi}
=
\frac{\partial\mathcal{L}_{\mathrm{ALCL}}}{\partial\sigma}
\frac{\partial\sigma}{\partial\tilde{\sigma}}
\frac{\partial\tilde{\sigma}}{\partial\psi}.
\label{eq24}
\end{equation}

\medskip

\noindent Thus, the residual-dependent ALCL objective provides gradients for both the network parameters and the adaptive loss parameters, coupling representation learning with adaptation of the loss geometry.

\subsection{$\psi$-Function Analysis and Robustness Properties of ALCL}
\label{subsec:influence_function}

A standard way to characterize the robustness of a loss function is through its $\psi$-function, which quantifies the sensitivity of the resulting estimation procedure to perturbations or contamination in the observations \cite{krasker1982efficient}. For the element-wise ALCL penalty, we characterize this behavior through its derivative with respect to the residual, $\psi(e)\triangleq\partial\rho(e)/\partial e$.

\begin{equation}
\rho(e;\alpha,\sigma) \;=\; \ln\!\left(1+\left(\frac{|e|}{\sigma}\right)^{\alpha}\right),
\qquad \alpha>1,\;\sigma>0.
\end{equation}

\medskip

\noindent Differentiating with respect to $e$ yields the ALCL $\psi$-function
\begin{align}
\psi_{\text{ALCL}}(e;\alpha,\sigma)
&= \frac{\partial}{\partial e}\ln\!\left(1+\left(\frac{|e|}{\sigma}\right)^{\alpha}\right) \nonumber\\
&= \frac{\alpha |e|^{\alpha-1}}{\sigma^{\alpha}+|e|^{\alpha}}\,\mathrm{sign}(e),
\label{eq:alcl_psi}
\end{align}

\noindent where $\mathrm{sign}(\cdot)$ denotes the sign operator. Geometrically, the score function $\psi_{\text{ALCL}}(e;\alpha,\sigma)$ describes how the loss surface responds to residual perturbations. The parameters $\alpha$ and $\sigma$ jointly control the curvature of the loss landscape: $\sigma$ determines the scale of the high-sensitivity region near the origin, while $\alpha$ modulates how rapidly the gradient attenuates as residual magnitude increases.

\subsubsection{Bounded influence and redescending behavior}

Let $u=|e|/\sigma\ge 0$. The magnitude of \eqref{eq:alcl_psi} can be written as

\begin{equation}
|\psi_{\text{ALCL}}(e;\alpha,\sigma)|
= \frac{1}{\sigma}\,\frac{\alpha u^{\alpha-1}}{1+u^{\alpha}}.
\label{eq:alcl_psi_u}
\end{equation}

\noindent Maximizing the right-hand side over $u\ge 0$ gives a unique maximizer

\begin{equation}
u^{\star} = (\alpha-1)^{1/\alpha}
\quad \Longleftrightarrow \quad
|e^{\star}|=\sigma(\alpha-1)^{1/\alpha},
\end{equation}

\noindent and the corresponding global bound

\begin{equation}
\sup_{e\in\mathbb{R}} |\psi_{\text{ALCL}}(e;\alpha,\sigma)|
= \frac{1}{\sigma}(\alpha-1)^{(\alpha-1)/\alpha} \;<\;\infty.
\label{eq:alcl_bounded_influence}
\end{equation}

\noindent The existence of a finite bound implies that the influence of any single observation is limited, preventing extreme residuals from dominating gradient updates. In geometric terms, the loss surface forms a broad basin around the origin where informative gradients are preserved, while the gradient magnitude reaches a maximum and subsequently decreases beyond a characteristic residual scale determined by $\alpha$ and $\sigma$. Moreover, as $|e|\to\infty$ we have

\begin{equation}
\psi_{\text{ALCL}}(e;\alpha,\sigma)
= \frac{\alpha}{|e|}\,\mathrm{sign}(e) + o\!\left(\frac{1}{|e|}\right),
\label{eq:alcl_bounded_influence_2}
\end{equation}

\noindent indicating that the $\psi$-function redescends to zero. This formally explains why ALCL suppresses extreme outliers, as very large residuals yield vanishingly small gradient contributions. Consequently, extremely large residuals exert diminishing influence on reconstruction-gradient contributions, reducing their potential to dominate network parameter updates.

\subsubsection{Comparison of $\psi$-functions}

\begin{figure}[h]
    \centering
    \includegraphics[width=\linewidth]{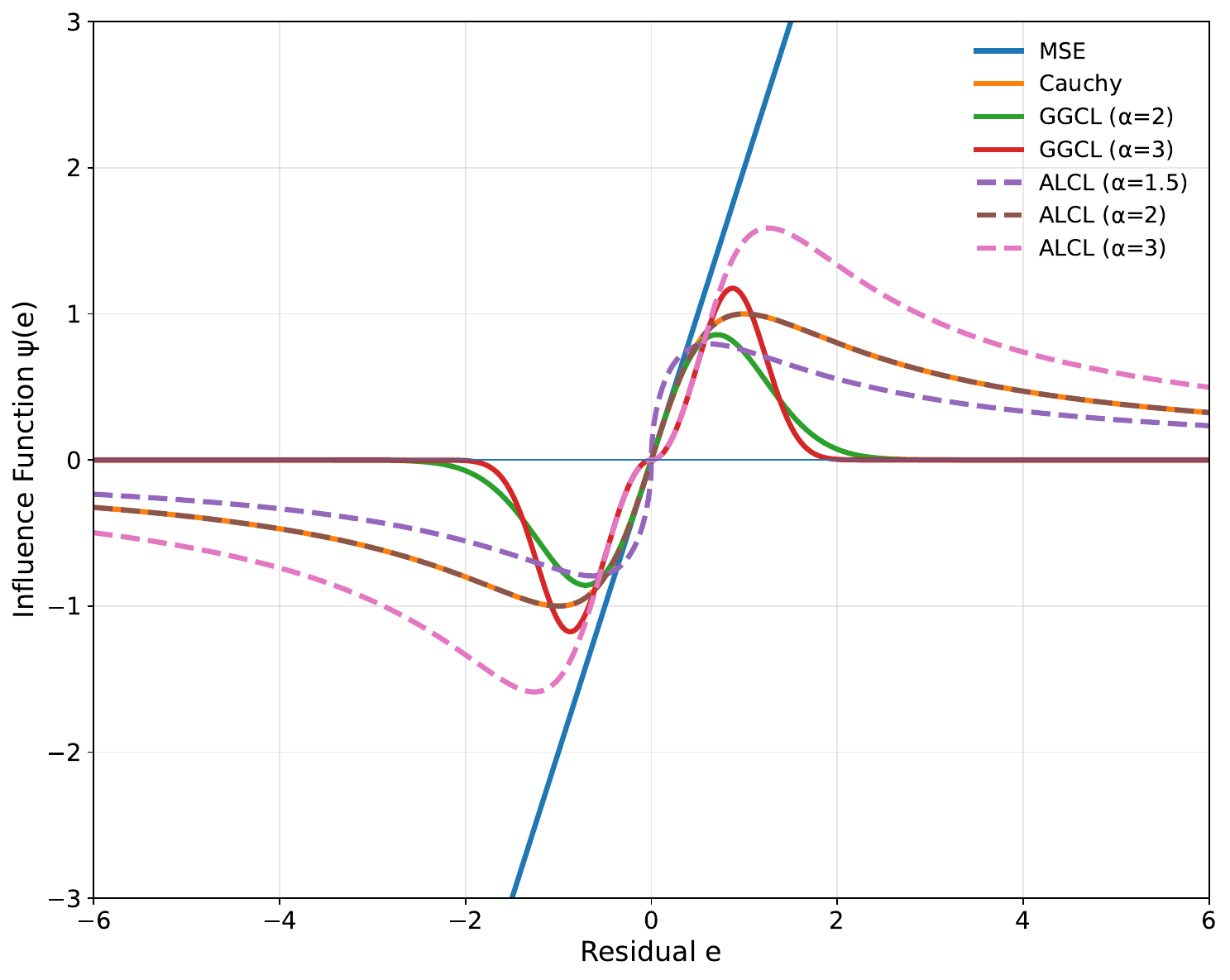}
    \caption{Comparison of the $\psi$-functions of MSE, Cauchy, GGCL, and ALCL for representative shape parameters under fixed scale settings. MSE exhibits unbounded linear growth, whereas the robust losses exhibit bounded and redescending behavior. ALCL allows its influence profile to adapt through jointly learned shape and scale parameters.}
    \label{fig1}
\end{figure}

\begin{figure}[h]
    \centering
    \includegraphics[width=\linewidth]{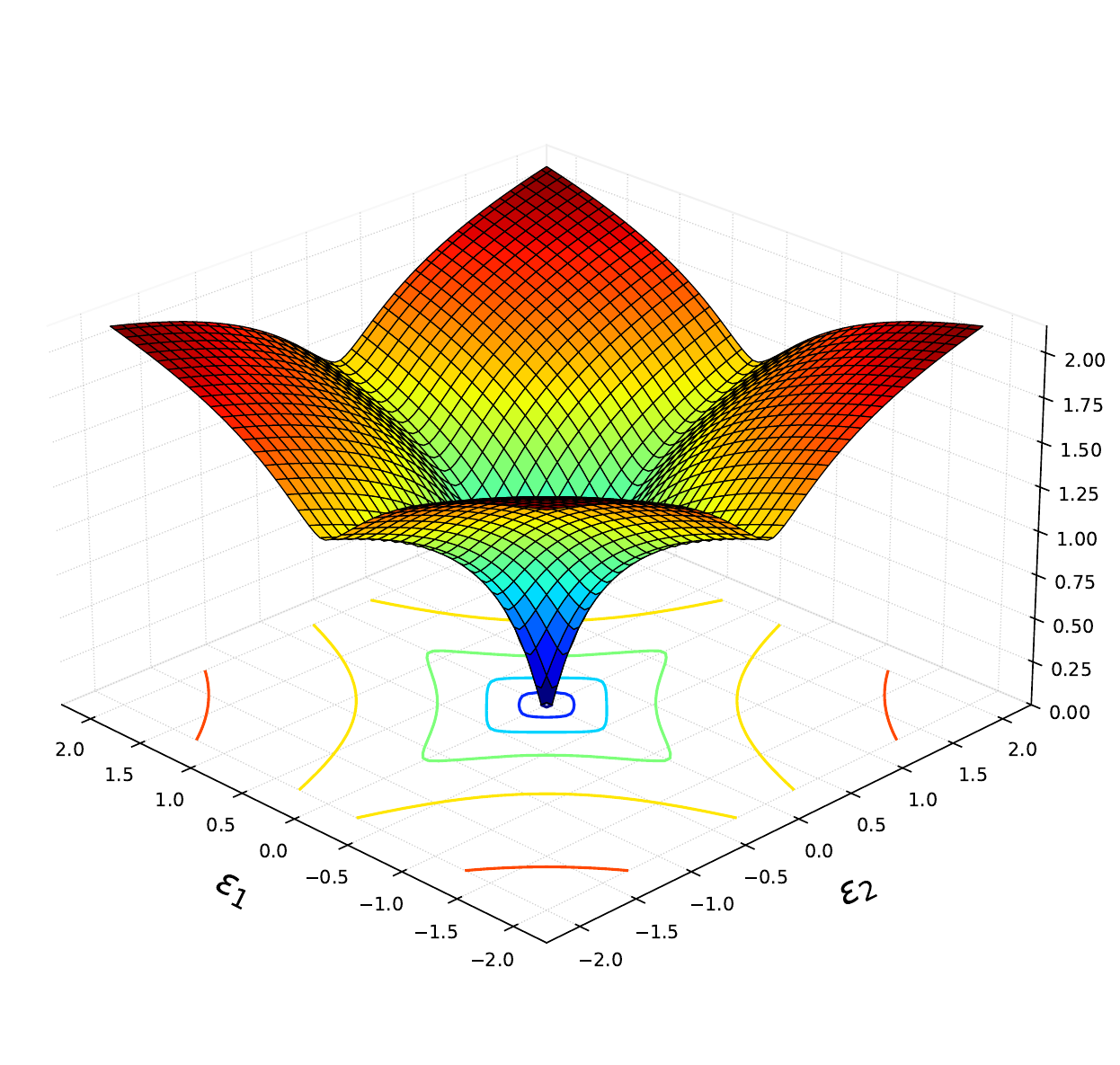}
    \caption{Two-dimensional loss geometry of ALCL in the residual space $(\epsilon_1,\epsilon_2)$. The surface exhibits a steep increase near the origin and a progressively decreasing slope at larger residual magnitudes, illustrating the redescending gradient behavior of ALCL.}
    \label{fig1_2}
\end{figure}

Equations \eqref{eq:alcl_bounded_influence} and \eqref{eq:alcl_bounded_influence_2} establish the bounded and redescending behavior of the ALCL $\psi$-function, respectively. Since $\alpha$ and $\sigma$ are learned jointly during training, the resulting influence profile can adapt as these parameters evolve. Figure~\ref{fig1} compares the $\psi$-functions of MSE, Cauchy, GGCL, and ALCL under identical scale settings. MSE assigns increasing gradient magnitude to larger residuals, making it sensitive to extreme deviations. In contrast, Cauchy, GGCL, and ALCL exhibit bounded and redescending $\psi$-functions, reducing the contribution of large residuals. For ALCL, variations in $\alpha$ modify the peak magnitude, location, and attenuation of the $\psi$-function, enabling adjustable sensitivity to moderate and extreme residuals. The plotted curves represent selected parameter settings, whereas $\alpha$ and $\sigma$ are learned during training. Figure~\ref{fig1_2} provides a complementary geometric visualization of ALCL. The loss surface increases more rapidly for small residuals and exhibits progressively reduced sensitivity as residual magnitudes grow. This behavior is consistent with the bounded and redescending residual gradients characterized by \eqref{eq:alcl_bounded_influence} and \eqref{eq:alcl_bounded_influence_2}.
 
\subsection{Regularization for ALCL}

For a batch of $N$ samples, the ALCL loss is defined as the average of the element-wise loss aggregated over spatial and channel dimensions

\begin{equation}
    \mathcal{L}_{\mathrm{ALCL}} = \frac{1}{N}\sum_{i=1}^{N}\frac{1}{|\Omega|}\sum_{j \in \Omega}\ln\!\left(1 + \left(\frac{|e_{ij}|}{\sigma_j}\right)^{\alpha}\right).
\label{eq27}
\end{equation}

\noindent In \eqref{eq27}, $\Omega$ indexes spatial and channel locations and $e_{ij}$ denotes the reconstruction residual at location $j$ for sample $i$. To further stabilize training and promote the correction of sparse impulsive noise, \eqref{eq28} shows the final loss incorporating a lightweight $\ell_1$ regularization term on the residuals

\begin{equation}
    \mathcal{L}_{\mathrm{total}} = \mathcal{L}_{\mathrm{ALCL}} + \lambda_{\ell_1}\,\mathbb{E}\!\left[|e|\right],
\label{eq28}
\end{equation}

\noindent where $\lambda_{\ell_1}$ modulates the degree of residual sparsity.

\section{ALCL-based autoencoder architectures}\label{sec4}

We introduce two architectural realizations of the proposed framework: a stacked autoencoder with ALCL loss (ALCL-SAE) for grayscale benchmarks and a convolutional autoencoder with ALCL loss (ALCL-CAE) for RGB natural images. Both architectures are designed to jointly optimize feature representations and the adaptive robustness parameters of the proposed loss.

\subsection{ALCL-SAE architecture for grayscale data}

The ALCL-SAE adopts a symmetric encoder--decoder structure to extract robust latent representations $\mathbf{z}$ from flattened noisy inputs $\mathbf{x}_{\text{noisy}} \in \mathbb{R}^{784}$~\cite{chen2016efficient,chen2018generalized,vincent2008extracting}. The encoder $f_{\theta_e}$ utilizes a sequence of dense layers (512--256--128), where each hidden layer $l$ produces a feature vector $\mathbf{h}^{(l)}$ as

\begin{equation}
    \mathbf{h}^{(l)} = \phi\!\left(\mathrm{BN}\!\left(\mathbf{W}^{(l)} \mathbf{h}^{(l-1)} + \mathbf{b}^{(l)}\right)\right),
\label{eq29}
\end{equation}

\noindent where $\mathbf{h}^{(l-1)}$ is the activation vector from the previous layer (with $\mathbf{h}^{(0)} = \mathbf{x}_{\text{noisy}}$), $\mathbf{W}^{(l)}$ and $\mathbf{b}^{(l)}$ denote the weights and bias terms, respectively, $\mathrm{BN}(\cdot)$ is batch normalization, and $\phi(\cdot)$ is the LeakyReLU activation. The final encoder layer yields the latent code $\mathbf{z} = \mathbf{h}^{(L)}$. The decoder $g_{\theta_d}$ mirrors this structure (256--512--784) and terminates with a Sigmoid activation to produce the reconstruction $\hat{\mathbf{x}} \in [0,1]^{784}$.

\subsection{ALCL-CAE architecture for RGB data}

For high-dimensional RGB data, the ALCL-CAE preserves spatial hierarchies through convolutional layers~\cite{wang2019symmetric,huang2023center}. The encoder $f_{\theta_e}$ maps the input tensor $\mathbf{H}^{(0)} \in \mathbb{R}^{32 \times 32 \times 3}$ to a bottleneck volume $\mathbf{Z} \in \mathbb{R}^{4 \times 4 \times 256}$ through three stages of $3 \times 3$ convolutions (64--128--256). The corresponding feature transformation at layer $l$ is given as

\begin{equation}
    \mathbf{H}^{(l)} = \phi\!\left(\mathrm{BN}\!\left(\mathbf{K}^{(l)} * \mathbf{H}^{(l-1)} + \mathbf{b}^{(l)}\right)\right),
\label{eq30}
\end{equation}

\noindent where $\mathbf{H}^{(l-1)}$ is the input feature map, $\mathbf{K}^{(l)}$ represents the convolutional kernels, and $*$ denotes the convolution operation. Spatial downsampling via max-pooling is applied after each block. The decoder $g_{\theta_d}$ progressively restores spatial resolution using convolutional layers interleaved with upsampling operations, culminating in a final Sigmoid-activated $3 \times 3$ convolutional layer to reconstruct the RGB image $\hat{\mathbf{X}} \in [0,1]^{32 \times 32 \times 3}$.

\subsection{Regularization and hyperparameter configuration}

To regularize network learning during the joint optimization of $\{\theta,\alpha,\sigma\}$, we incorporate a residual-based $\ell_1$ penalty and an $\ell_2$ penalty on selected network weight matrices. The total training loss is defined as

\begin{equation}
    \mathcal{L}_{\text{total}}
    =
    \mathcal{L}_{\text{ALCL}}
    +
    \lambda_{\ell_1}\mathbb{E}[|e|]
    +
    \zeta\|\mathbf{W}\|_2^2,
\label{eq31}
\end{equation}

\noindent where the residual $\ell_1$ regularization weight $\lambda_{\ell_1}$ is fixed at $10^{-4}$ for all experiments, and $\mathbf{W}$ denotes the network weight matrices subject to $\ell_2$ regularization. The regularization coefficient $\zeta$ is set to $10^{-2}$ for the ALCL-SAE and $10^{-4}$ for the ALCL-CAE. The $\ell_2$ penalty is incorporated into the training objective through layer-specific kernel regularization. Preliminary hyperparameter evaluations indicated that $\ell_2$ regularization did not improve the performance of the MSE and GGCL baselines; therefore, these models were trained without this penalty in the reported experiments.

\subsection{Channel-wise adaptive scaling}

For the multi-channel ALCL-CAE, we extend the scale parameter to a vector$\boldsymbol{\sigma}\in[\sigma_{\min},\sigma_{\max}]^C$, where $C$ denotes the number of image channels and $\sigma_{\min}>0$ and $\sigma_{\max}$ define the lower and upper bounds, respectively. Each channel is assigned an independently trainable scale parameter $\sigma_c$, enabling ALCL to accommodate channel-dependent variations in reconstruction residuals. The resulting channel-wise objective is defined as
\begin{equation}
    \mathcal{L}_{\text{ALCL}}(E; \alpha, \boldsymbol{\sigma})
    = \frac{1}{HWC}
    \sum_{c=1}^{C} \sum_{h=1}^{H} \sum_{w=1}^{W}
    \ln\!\left(
    1 + \left(\frac{|E_{h,w,c}|}{\sigma_c}\right)^{\alpha}
    \right),
    \label{eq32}
\end{equation}

\noindent where $E_{h,w,c}$ denotes the reconstruction residual at spatial coordinates $(h,w)$ for channel $c$, and $H$ and $W$ represent the image height and width, respectively.The channel-specific scale parameters are optimized jointly with the network parameters through backpropagation,allowing the loss to adjust its sensitivity to residuals across different channels while maintaining a consistent scale parameter within each channel.

\section{Experimental setup}\label{sec5}

\subsection{Dataset}

To evaluate the robustness and generalizability of the proposed ALCL, we utilize four widely used image classification benchmarks spanning different dimensionalities and visual complexities: MNIST~\cite{Lecun1998}, Fashion-MNIST~\cite{xiao2017fashion}, CIFAR-10~\cite{krizhevsky2009learning}, and Street View House Numbers (SVHN)~\cite{netzer2011reading}. MNIST consists of 70,000 single-channel grayscale images ($28\times28$ pixels) across 10 digit classes (60,000 training, 10,000 test). Fashion-MNIST shares the same data format and split size ($28\times28$ grayscale, 60,000 training, 10,000 test), but its 10 classes represent diverse clothing categories, introducing greater intra-class appearance variability. CIFAR-10 contains 60,000 RGB natural images ($32\times32\times3$ pixels) across 10 object classes (50,000 training, 10,000 test) with substantial variations in color, texture, and background structure. Similarly, SVHN comprises $32\times32\times3$ RGB images of house numbers from real-world scenes, consisting of 73,257 training and 26,032 test samples.

For all datasets, pixel intensities were normalized to $[0,1]$ prior to noise injection. For model validation, 10\% of the original training data was separated via stratified sampling with a fixed random seed to form a validation set, preserving class distributions. This yielded final partitions of 54,000 training, 6,000 validation, and 10,000 test samples for MNIST and Fashion-MNIST; 45,000 training, 5,000 validation, and 10,000 test samples for CIFAR-10; and 65,931 training, 7,326 validation, and 26,032 test samples for SVHN. Identical partitions, preprocessing protocols, and noise settings were applied across ALCL, MSE, and GGCL baselines to ensure a direct and consistent comparison. Together, these datasets provide complementary evaluation settings ranging from relatively simple grayscale digits and clothing images to more structurally complex natural and real-world RGB imagery.

\subsection{Non-Gaussian mixed heavy-tailed noise}

To evaluate robustness under non-Gaussian noise, we adopt a mixed heavy-tailed noise model that combines dense background noise with sparse impulsive spikes. The background component follows a Cauchy distribution, defined by the probability density function

\begin{equation}
    f(x; 0, \gamma) = \frac{1}{\pi \gamma \left[1 + (x/\gamma)^2\right]} .
\label{eq33}
\end{equation}

\noindent In \eqref{eq33}, the scale parameter $\gamma$ controls the dispersion of the distribution and directly influences the magnitude of heavy-tailed deviations. Impulsive noise is added to model localized, high-magnitude interference as

\begin{equation}
    \mathbf{N}_{\text{imp}} = \mathbf{B} \cdot \xi ,
\label{eq34}
\end{equation}

\noindent where $\mathbf{B} \sim \text{Bernoulli}(P)$ selects a fraction $P$ of noisy locations, and $\xi \in \{-M, M\}$ denotes the impulse magnitude. Table~\ref{tab1} lists the values of $\gamma$, $P$, and $M$ defining the low- and high-noise settings used in the experiments. The noise parameters in Table \ref{tab1} are defined relative to each data modality, as identical parameter values do not represent uniform severity across grayscale and RGB inputs. Due to the higher dimensionality and per-channel perturbations inherent in RGB data, a specific noise setting induces substantially stronger distortions than it would in a grayscale input. For grayscale benchmarks, the low-noise regime approximates a near-Gaussian distribution, resulting in marginal performance gaps between loss functions. In contrast, these same settings produce aggressive non-Gaussian corruptions in RGB datasets. Consequently, noise intensities were calibrated to ensure that the corrupted signals remain within a range where semantic features are preserved for reconstruction, avoiding a total collapse of the underlying data structure.

\begin{table}[h]
\caption{Mixed heavy-tailed noise settings used in the experiments.}\label{tab1}%
\resizebox{\columnwidth}{!}{%
\begin{tabular}{@{}lllll@{}}
\toprule
\textbf{Dataset Type} & \textbf{Noise Level} & $\boldsymbol{\gamma}$ & $\boldsymbol{P}$ & $\boldsymbol{M}$ \\
\midrule
Grayscale (MNIST, Fashion-MNIST)    & High & 1.5   & 0.20  & 0.5 \\
                                    & Low  & 0.2   & 0.02  & 0.5 \\
\hline
RGB (CIFAR-10, SVHN)                & High & 0.2   & 0.02  & 0.5 \\
                                    & Low  & 0.02  & 0.002 & 0.05 \\
\bottomrule
\end{tabular}
}
\end{table}

\subsection{Evaluation protocol and classifier architectures}

\begin{figure*}[t]
\centering
\includegraphics[width=\textwidth]{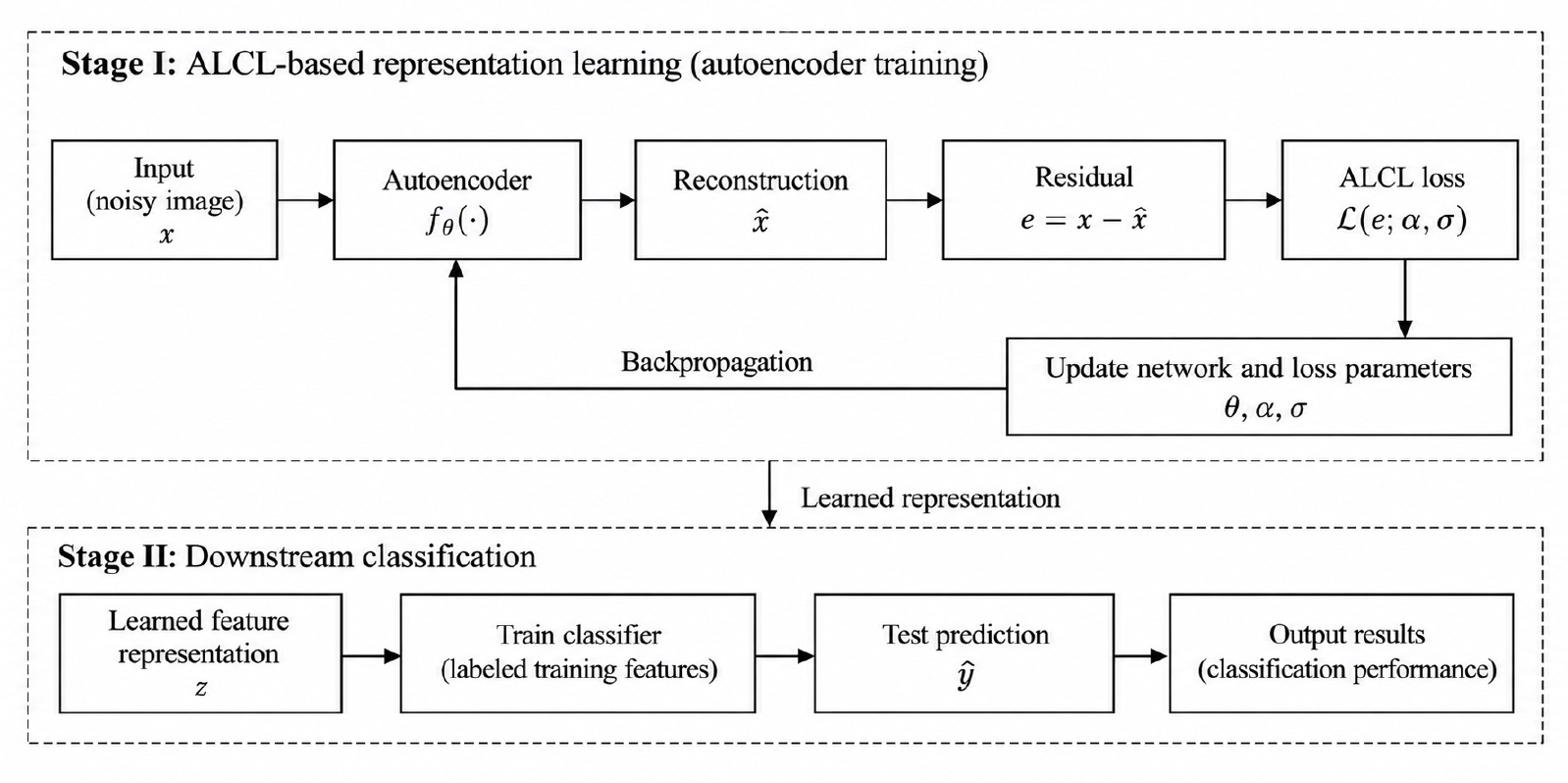}
\caption{Learning workflow of the proposed ALCL framework, including joint adaptive optimization during autoencoder training and subsequent downstream classification.}
\label{fig:ALCL_flowchart}
\end{figure*}

To quantify the robustness of the learned representations, we employ a downstream classification task operating directly on the latent features extracted from the autoencoder bottleneck. In the proposed framework, for the MNIST and Fashion-MNIST datasets, the bottleneck features are fed into a deep multilayer perceptron classifier consisting of three hidden dense layers with 256, 128, and 64 neurons, respectively. Each hidden layer uses LeakyReLU activation and Dropout with probabilities ranging from 0.3 to 0.4 to promote regularization and training stability. For the CIFAR-10 and SVHN datasets, we adopt a convolutional classifier to better exploit spatial structure. This model comprises two $3 \times 3$ convolutional layers with 256 and 128 filters, respectively, followed by global average pooling to reduce the spatial dimensions into a feature vector. A subsequent dense layer with 256 units and a softmax output layer perform the final classification. Figure~\ref{fig:ALCL_flowchart} illustrates the complete learning workflow of the proposed ALCL framework, including joint optimization of the network and adaptive robustness parameters during autoencoder training, followed by learned representation extraction, classifier training, and test prediction.

\section{Experimental results and discussion}\label{sec6}

In this study, we compare ALCL against MSE and the GGCL to evaluate robustness under non-Gaussian noise on grayscale and RGB benchmarks. MSE quantifies performance degradation when training is exposed to heavy-tailed and impulsive noise. GGCL generalizes correntropy by adopting a kernel derived from the GCD, allowing it to emulate Gaussian, Laplacian, or Cauchy kernel through fixed choices of shape and scale parameters. All results are compared using identical architectures and training protocols to ensure a fair comparison.

\subsection{Visual reconstruction analysis}

\begin{figure*}[h]
\centering
\includegraphics[width=\linewidth]{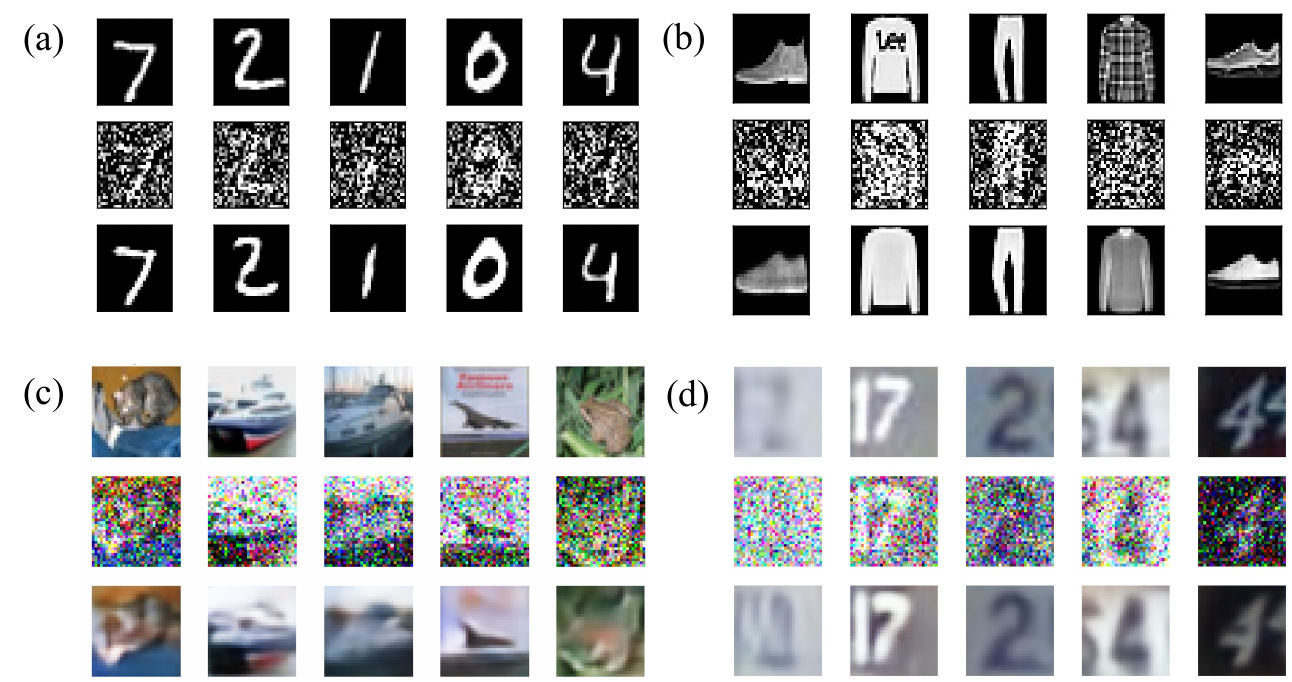}
\caption{ Qualitative reconstruction results under non-Gaussian mixed noise. For each dataset, the rows show the clean input image (top), the input corrupted by high-level heavy-tailed and impulsive noise (middle), and the reconstructed output obtained using the proposed ALCL framework (bottom). Results are shown for (a) MNIST, (b) Fashion-MNIST, (c) CIFAR-10, and (d) SVHN.}
\label{fig2}
\end{figure*}

The visual results in Figure~\ref{fig2} demonstrate that ALCL effectively restores the signal even when substantial noise obscures the original structures. For MNIST and Fashion-MNIST, the reconstructed outputs successfully recover the sharp edges necessary for classification. In the RGB datasets, CIFAR-10 and SVHN, ALCL removes impulsive artifacts while preserving color and structural integrity. This confirms that rather than simply smoothing the data, the proposed framework adaptively suppresses outliers to facilitate a robust reconstruction of the original input.

\begin{table}[h]
\caption{Classification accuracy (\%) comparison under high and low non-Gaussian noise conditions (mean $\pm$ standard deviation).}\label{tab2}%
\resizebox{\columnwidth}{!}{%
\begin{tabular}{@{}lllll@{}}
\toprule
\textbf{Dataset} & \textbf{Noise Level} & \textbf{MSE} & \textbf{GGCL} & \textbf{ALCL} \\
\midrule
MNIST         & High & $83.45 \pm 0.23$ & $85.82 \pm 0.53$ & $\mathbf{88.20 \pm 0.23}$ \\
              & Low  & $97.27 \pm 0.06$ & $97.29 \pm 0.08$ & $\mathbf{97.74 \pm 0.08}$ \\ \hline
Fashion-MNIST & High & $71.73 \pm 0.40$ & $72.24 \pm 0.08$ & $\mathbf{74.77 \pm 0.28}$ \\
              & Low  & $85.40 \pm 0.20$ & $85.77 \pm 0.15$ & $\mathbf{85.86 \pm 0.10}$ \\ \hline
CIFAR-10      & High & $54.07 \pm 0.18$ & $57.91 \pm 0.34$ & $\mathbf{58.58 \pm 0.337}$ \\
              & Low  & $66.38 \pm 0.10$ & $69.30 \pm 0.13$ & $\mathbf{70.03 \pm 0.42}$ \\ \hline
SVHN          & High & $71.50 \pm 0.16$ & $72.15 \pm 0.39$ & $\mathbf{73.66 \pm 0.19}$ \\
              & Low  & $85.56 \pm 0.37$ & $85.86 \pm 0.21$ & $\mathbf{85.98 \pm 0.15}$ \\
\bottomrule
\end{tabular}
}
\end{table}

\subsection{Performance comparison on benchmarks}

Table~\ref{tab2} summarizes the average classification accuracy across all benchmarks. The results demonstrate that ALCL consistently outperforms both MSE and GGCL in every noise scenario. Under high noise, ALCL exhibits its largest performance gains. On grayscale benchmarks, MNIST and Fashion-MNIST, ALCL achieves up to 4.75\% higher accuracy than MSE. For complex RGB datasets, ALCL reaches 58.58\% on CIFAR-10 and 73.66\% on SVHN, surpassing the exhaustively tuned GGCL. This improvement is consistent with the bounded and redescending influence of ALCL, which limits the contribution of large residuals during optimization. For low noise, the performance gaps narrow as the corruption becomes less severe, although ALCL maintains the highest mean classification accuracy across all four datasets.

\begin{figure}[h]
\centering
\includegraphics[width=\linewidth]{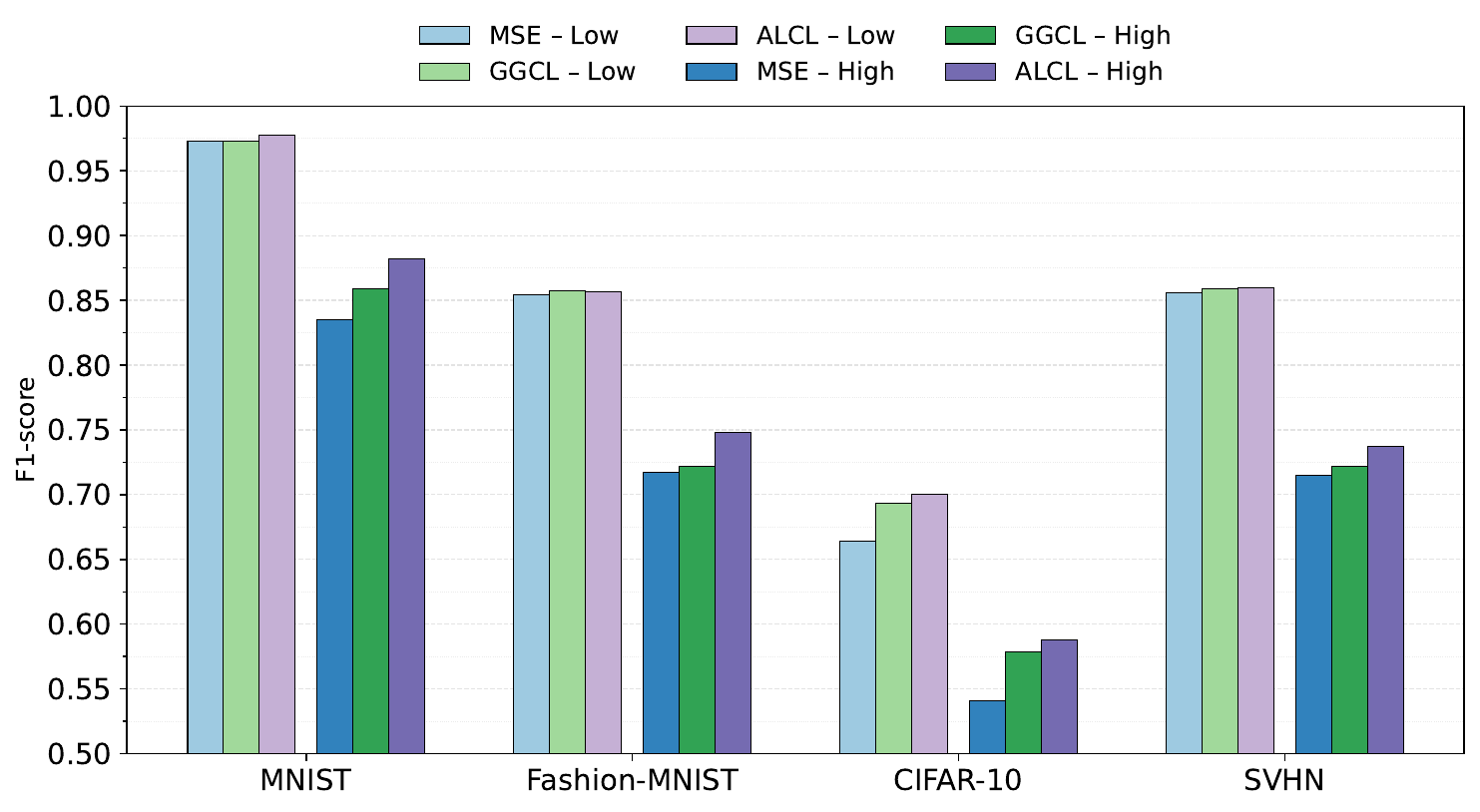}
\caption{Comparison of F1-score for MSE, GGCL, and ALCL under high- and low-noise conditions across the four benchmark
datasets.}
\label{fig:ALCL_Low_High_F1}
\end{figure}

The F1-score comparison further supports the robustness of ALCL across noise levels. Under high non-Gaussian noise, ALCL achieves the highest F1-score across all four datasets, whereas the differences generally become smaller under low noise, as shown in Fig.~\ref{fig:ALCL_Low_High_F1}. GGCL provides a strong robust baseline when its parameters are carefully tuned; however, its loss parameters remain fixed during training. In contrast, the jointly learned shape and scale parameters of ALCL allow its robustness characteristics to adapt during optimization, with the resulting advantage being most evident under stronger non-Gaussian corruption.

\subsection{Convergence behavior}

\begin{figure}[h]
\centering
\includegraphics[width=\linewidth]{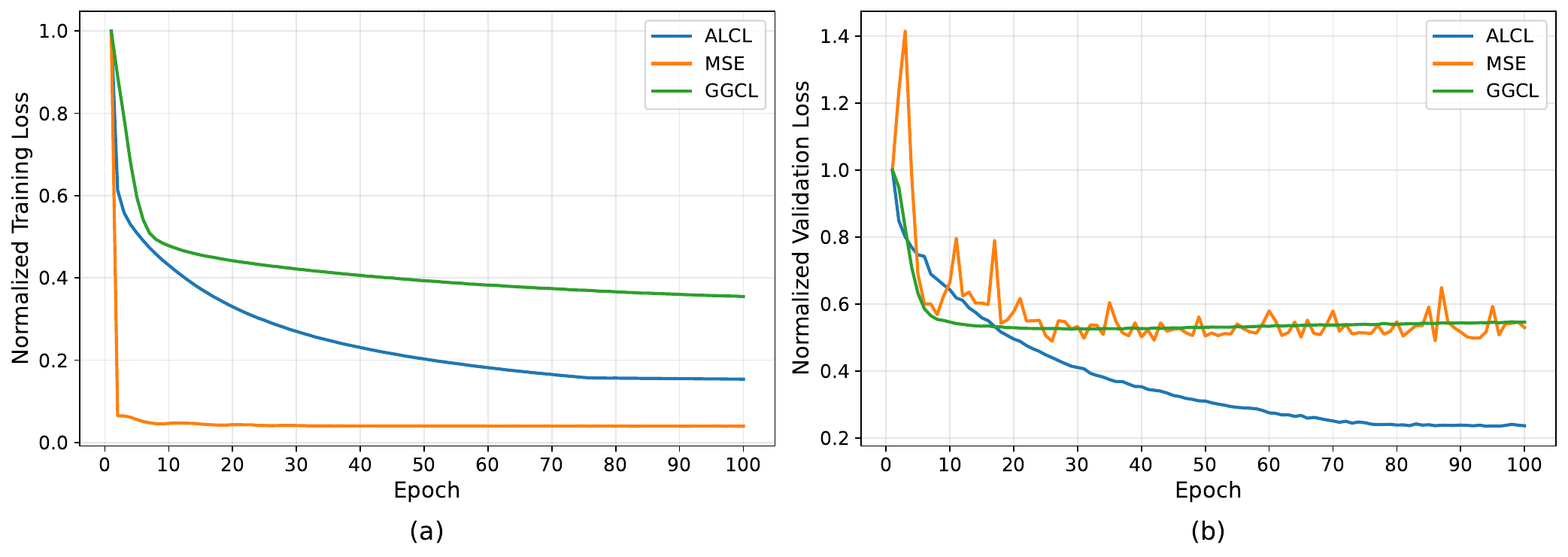}
\caption{Normalized training and validation loss trajectories for MSE, GGCL, and ALCL. Each trajectory is normalized by its corresponding first-epoch loss to enable comparison of relative convergence behavior across the different loss functions: (a) training loss and (b) validation loss.}
\label{fig:Normalized_Training_Validation_Loss}
\end{figure}

To further examine the optimization behavior of the competing loss functions, we analyze their training and validation loss trajectories over successive epochs. Since ALCL, MSE, and GGCL optimize objectives with different numerical scales, each trajectory is normalized by its corresponding first-epoch value to enable comparison of relative convergence behavior, as shown in Fig.~\ref{fig:Normalized_Training_Validation_Loss}. While all methods show an initial reduction in training loss, ALCL continues to decrease progressively across subsequent epochs. A similar trend is observed for validation loss, where MSE and GGCL stabilize earlier, whereas ALCL exhibits sustained reduction without evident divergence between training and validation behavior. These results indicate a sustained reduction in the normalized ALCL objective without evident training–validation divergence and complement its downstream performance improvements.

\subsection{Statistical stability and reliability}

\begin{figure}[h]
\centering
\includegraphics[width=\linewidth]{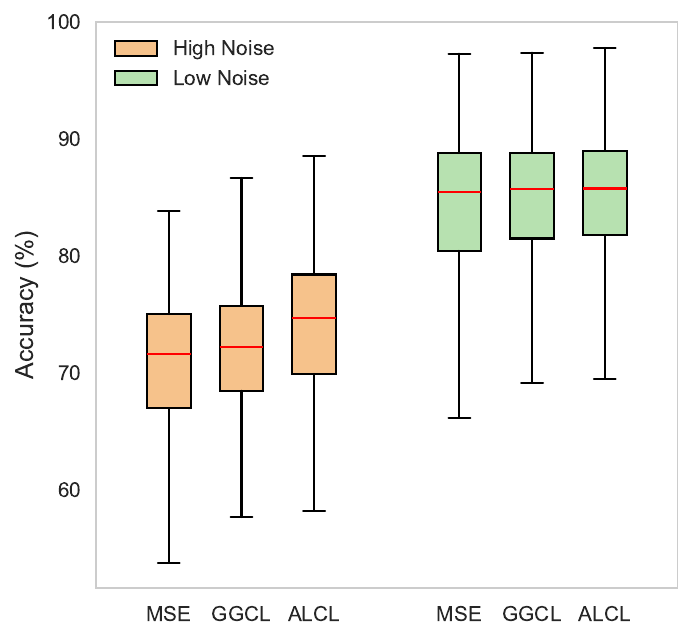}
\caption{Boxplot of classification accuracy for MSE, GGCL, and ALCL under high and low noise conditions. ALCL achieves higher median accuracy and exhibits tighter interquartile ranges across noise regimes, indicating superior robustness and statistical stability.}
\label{boxplotall}
\end{figure}

Figure~\ref{boxplotall} illustrates the classification accuracy distributions across datasets under high- and low-noise conditions. Under high noise, ALCL shows higher median accuracy than MSE and GGCL, whereas the differences become smaller under low noise. The relatively compact distributions further indicate stable performance across repeated runs, complementing the mean and standard deviation results reported in Table~\ref{tab2}.

\begin{table}[h]
\caption{Paired two-tailed t-test p-values (n=5) comparing ALCL against MSE and GGCL under high- and low-noise conditions. Statistical significance is determined at the 0.05 level. Bold values indicate non-significant differences ($p \geq 0.05$).}\label{tab3}%
\resizebox{\columnwidth}{!}{%
\begin{tabular}{@{}llll@{}}
\toprule
\textbf{Dataset} & \textbf{Noise} & \textbf{ALCL vs MSE} & \textbf{ALCL vs GGCL} \\
\midrule
MNIST           & High & $1.1\times10^{-6}$ & 0.0023 \\
                & Low  & 0.0008             & 0.0005 \\ \hline
Fashion-MNIST   & High & 0.0002             & $3.5\times10^{-5}$ \\
                & Low  & 0.015              & \textbf{0.58} \\ \hline
CIFAR-10        & High & $1.6\times10^{-6}$ & \textbf{0.13} \\
                & Low  & $1.0\times10^{-6}$ & 0.028 \\ \hline
SVHN            & High & $6.5\times10^{-5}$ & 0.0018 \\
                & Low  & \textbf{0.066} & \textbf{0.12} \\
\bottomrule
\end{tabular}
}
\end{table}

To examine the statistical significance of the results, paired two-tailed t-tests (n=5) were conducted for each dataset and noise condition, with p-values reported in Table~\ref{tab3}. Under high-noise conditions, ALCL achieves statistically significant improvements over MSE across all datasets and over GGCL in three out of four cases, confirming the effectiveness of its adaptive heavy-tailed modeling when residual distributions deviate from Gaussian assumptions. In low-noise conditions, the performance differences are smaller. Nevertheless, ALCL maintains statistically significant improvements over MSE in three datasets, while exhibiting performance comparable to GGCL in certain cases. The statistical results in Table~\ref{tab3} support the stability trends observed in Fig.~\ref{boxplotall}, demonstrating that ALCL provides strong robustness under high-noise conditions while maintaining competitive performance when the noise level is low and the residual distribution is close to Gaussian.

\section{Conclusion}\label{sec7}

This study introduces the Adaptive Log-Correntropy Loss (ALCL), a principled heavy-tailed loss for robust deep learning under non-Gaussian noise. By jointly learning the shape ($\alpha$) and scale ($\sigma$) parameters alongside the network weights through differentiable reparameterization, ALCL adapts its loss geometry during optimization, overcoming the limitations of robust losses with fixed parameters. Geometric and influence-function analyses demonstrate that ALCL providesbounded and redescending influence, preserving informative gradients for moderate residuals while suppressing the contribution of extreme outliers. The convergence analysis further shows a sustained reduction in normalized training and validation loss for ALCL, whereas the fixed-loss baselines exhibit earlier stabilization. Experiments on grayscale and RGB benchmarks under mixed heavy-tailed and impulsive noise show that ALCL consistently achieves higher mean classification accuracy than MSE and exhaustively tuned GGCL, with its advantages being most pronounced under high-noise conditions.

The present study establishes the effectiveness of ALCL under controlled heavy-tailed and impulsive corruption, while also identifying important questions for further investigation. In particular, the current experiments consider statistically defined noise conditions during training and do not explicitly examine how the jointly learned shape and scale parameters respond when the noise characteristics change abruptly or evolve over time. This behavior is particularly relevant to nonstationary time-series and communication signals, where sudden disturbances may alter the residual distribution during operation. The current evaluation focuses on reconstructing data corrupted by non-Gaussian noise, providing a foundation for investigating ALCL under more complex and domain-specific corruption conditions.  In other words, the current experimental setup is limited to controlled combinations of Cauchy-distributed background noise and sparse impulsive disturbances. Although these conditions provide a useful benchmark for heavy-tailed corruption, they do not fully represent the diversity of non-Gaussian noise encountered in practical applications, including asymmetric, multimodal, and temporally correlated disturbances.

Future work will therefore investigate the response and adaptation behavior of ALCL under dynamically changing and naturally occurring noise, with particular emphasis on multichannel time-series and communication data, as well as medical imaging applications affected by non-Gaussian corruption. These studies will examine how the learned loss geometry reacts to abrupt changes in the noise regime and whether robust performance is maintained across changing corruption conditions. Such future work can also help investigate the convergence, stability, and sensitivity of the learned shape and scale parameters, particularly under abrupt changes in noise characteristics and help provide further insight into how the adaptive loss geometry responds to evolving residual distributions to establish practical guidelines for parameter initialization and optimization. Finally, another important direction is to incorporate ALCL directly into classification objectives, extending its adaptive robustness mechanism from reconstruction to discriminative learning and evaluating its effectiveness as a robust classification loss under heavy-tailed and impulsive noise. 

\bibliographystyle{IEEEtran}
\bibliography{reference1.bib}

\end{document}